\documentclass[11pt]{article}
\usepackage{eamt20}
\usepackage{times}
\usepackage{url}
\usepackage{latexsym}
\usepackage[small,bf]{caption} 
\usepackage{microtype}

\usepackage[russian,english]{babel}
\usepackage[T2A,T1]{fontenc}
\usepackage[utf8]{inputenc}
\DeclareUnicodeCharacter{2588}{\rule{.3em}{.9em}}

\usepackage{url}
\usepackage{gnuplot-lua-tikz}
\usepackage{tikz}
\usepackage{pgfplots}
\usepackage[]{algorithm2e}
\usepackage[normalem]{ulem}
\usepackage{multirow}
\usepackage{rotating}
\usepackage{amsmath}
\usepackage[noabbrev,capitalise]{cleveref}
\usepackage{booktabs,tabularx}
\newcolumntype{Y}{>{\raggedleft\arraybackslash}X} 
\newcolumntype{C}{>{\centering\arraybackslash}X}

\def\rot#1{{\begin{sideways}#1\end{sideways}}}
\def\circled#1{\textcircled{\raisebox{-.9pt}{#1}}}

\def\splithyphen{\discretionary{-}{-}{-}}
\def\langpair#1#2{#1\splithyphen{}to\splithyphen{}#2}

\def\parcite#1{\cite{#1}} 
\def\perscite#1{\newcite{#1}} 
\def\significantmark{$\ddagger$}
\def\significant{\rlap{ $\ddagger$}} 

\newcommand{\COtwo}{$\text{CO}_2$}

\usepackage{array}
\newcolumntype{H}{>{\setbox0=\hbox\bgroup}c<{\egroup}@{}}

\title{Efficiently Reusing Old Models Across Languages via Transfer Learning}

\author{Tom Kocmi
        \qquad Ond{\v{r}}ej Bojar
		\\ \\
        Charles University, Faculty of Mathematics and Physics \\
        Institute of Formal and Applied Linguistics \\
        Malostransk{\'{e}} n{\'{a}}m{\v{e}}st{\'{\i}} 25, 118 00 Prague, Czech Republic \\
         {\tt \{kocmi,bojar\}@ufal.mff.cuni.cz}}

\date{}

\begin{document}
\maketitle
\begin{abstract}

Recent progress in neural machine translation is directed towards larger neural networks trained on an increasing amount of hardware resources.
As a result, NMT models are costly to train, both financially, due to the electricity and hardware cost, and environmentally, due to the carbon footprint.
It is especially true in transfer learning for its additional cost of training
the ``parent'' model before transferring knowledge and training the desired ``child'' model.
In this paper, we propose a simple method of re-using an already trained model for different language pairs where there is no need for modifications in model architecture.
Our approach does not need a separate parent model for each investigated language pair, as it is typical in NMT transfer learning. To show the applicability of our method, we recycle a Transformer model trained by different researchers and use it to seed models for different language pairs. 
We achieve better translation quality and shorter convergence times than when training from random initialization. 

\end{abstract}

\section{Introduction}

Neural machine translation (NMT), the current prevalent approach to automatic translation, is known to require large amounts of parallel training sentences and an extensive amount of training time on dedicated hardware. 
The total training time significantly increases, especially when training strong baselines, searching for best hyperparameters or training multiple models for various language pairs.

\perscite{schwartz2019green} analyzed 60 papers from top AI conferences and found out that 80\% of them target accuracy over efficiency, and only a small portion of papers argue for a new efficiency result. They also noted that the increasing financial cost of the computations could make it difficult for researchers to engage in deep learning research or limit training strong baselines. Furthermore, increased computational requirements have also an environmental cost.
\perscite{strubell2019energy} estimated that training a single Transformer ``big'' model produces 87 kg of \COtwo{} and that the massive Transformer architecture parameter search produced 298 tonnes of \COtwo{}.\footnote{The paper reports numbers based on the U.S. energy mix.}

However, a lot of research has been already invested into cutting down the long training time by the design of NMT model architectures, promoting self-attentive \parcite{vaswani:2017} or convolutional \parcite{gehring2017convolutional} over recurrent ones \parcite{bahdanau:2014:corr} or the implementation of heavily optimized toolkits \parcite{mariannmt}.

In this paper, we propose a novel view on re-using already trained ``parent'' models without the need to prepare a parent model in advance or modify its training hyper-parameters. 
Furthermore, we propose a second method based on a vocabulary transformation
technique that makes even larger improvements, especially for languages using an
alphabet different from the re-used parent model. Our transfer learning approach
leads to better performance as well as faster convergence speed of the
``child'' model compared to training the model from scratch. We document that
our methods are not restricted only to low-resource languages, but they can be used even for high-resource ones.

Previous transfer learning techniques \parcite{neubig-hu:2018:EMNLP,kocmi-adaption} rely on a shared vocabulary between the parent and child models. As a result, these techniques separately train parent model for each different child language pair. In contrast, our approach can re-use one parent model for multiple various language pairs, thus further lowering the total training time needed.

In order to document that our approach is not restricted to parent models trained by us, we re-use parent model trained by different researchers: we use the winning model of WMT 2019 for Czech-English language pair \parcite{popel-EtAl:2019:WMT}.

The paper is organized as follows: \cref{sec:method_desc} describes the method of Direct Transfer learning, including our improvement of vocabulary transformation.
\cref{sec:experiments} presents the model, training data, and our experimental setup.
\cref{sec:results} describes the results of our methods followed by the analysis in \cref{sec:analysis}. Related work is summarized in \cref{sec:related} and we conclude the discussion in \cref{sec:conclusion}.

\section{Transfer Learning}
\label{sec:method_desc}

In this work, we present the use of transfer learning to reduce the training time and improve the performance in comparison to training from random initialization even for high-resource language pairs.

Transfer learning is an approach of using training data from a related task to improve the accuracy of the main task in question \parcite{tan2018survey}. One of the first transfer learning techniques in NMT was proposed by \perscite{zoph-EtAl:2016:EMNLP2016}. They used word-level NMT and froze several model parts, especially embeddings of words that are shared between parent and child model.

We build upon the work of \perscite{kocmi-adaption}, who simplified the transfer learning technique thanks to the use of subword units \parcite{wu2016google} in contrast to word-level NMT transfer learning \parcite{zoph-EtAl:2016:EMNLP2016} and extended the applicability to unrelated languages.

Their only requirement, and also the main disadvantage of the method, is that the vocabulary has to be shared and constructed for the given parent and child languages jointly, which makes the parent model usable only for the particular child language pair. This substantially increases the overall training time needed to obtain the desired NMT system for the child language pair. 

The method of \perscite{kocmi-adaption} consists of three steps: (1) construct the vocabulary from both the parent and child corpora, (2) train the parent model with the shared vocabulary until convergence, and (3) continue training on the child training data. 

\perscite{neubig-hu:2018:EMNLP} call such approaches warm-start, where we use the child language pair to influence the parent model. In our work, we focus on the so-called cold-start scenario, where the parent model is trained without a need to know the language pair in advance. Therefore we cannot make any modifications of the parent training to better handle the child language pair. 
The cold-start transfer learning is expected to have slightly worse performance than the warm-start approach. However, it allows reusing one parent model for multiple child language pairs, which reduces the total training time in comparison to the use of warm-start transfer learning.

We present two approaches: Direct Transfer that ignores child-specific vocabulary altogether; and Transformed Vocabulary, which modifies vocabulary of the already trained parent. Thus, one parent model can be used for multiple child language pairs.

\subsection{Direct Transfer}

Direct Transfer can be seen as a simplification of \perscite{kocmi-adaption}. We ignore the specifics of the child vocabulary and train the child model using the parent vocabulary. We suppose that the subword vocabulary can handle the child language pair, although it is not optimized for it.

We take an already trained model and use it as initialization for a child model using a different language pair. We continue the training process without any change to the vocabulary or hyper-parameters. This applies even to the training parameters, such as the learning rate or moments.

This method of continued training on different data while preserving hyper-parameters is used under the name ``continued training'' or ``fine-tuning'' \parcite{hinton2006reducing,micelibarone2017EMNLP2017}, but it is mostly used as a domain adaptation within a given language pair.

Direct Transfer relies on the fact that the current NMT uses subword units instead of words.
The subwords are designed to handle unseen words or even characters, breaking the input into shorter units, possibly down to individual bytes as implemented, for example, by Tensor2Tensor \parcite{tensor2tensor}. This property ensures that the parent vocabulary can, in principle, serve for any child language pair, but it can be highly suboptimal, segmenting child words into too many subwords.

We present an example of a Russian phrase and its segmentation based on English-Czech or English-Russian vocabulary in \cref{tab:vocab_variant_usage}. When using child-specific vocabulary, the segmentation works as expected, splitting the phrase into three tokens. However, when we use a vocabulary that contains only the Cyrillic alphabet\footnote{This happened solely due to noise in the Czech-English parent training data.} and not many longer sequences of characters, the sentence is split into 13 tokens. We can notice that English-Czech wordpiece vocabulary is missing a character \foreignlanguage{russian}{``Л''}, thus it breaks it into the byte representation ``\verb|\1051;|''.

\begin{table}[t]
\centering
\begin{tabular}{l|cc|cc}
\multicolumn{1}{c}{} & \multicolumn{2}{c|}{Child-specific} & \multicolumn{2}{c}{EN-CS vocab.}  \\
\multicolumn{1}{r}{Avg. \# per:} & Sent. & Word & Sent. & Word \\
\hline               
Odia     & 95.8 & 3.7 & 496.8 & 19.1\\
Estonian & 26.0 & 1.1 &  56.2 &  2.3\\
Finnish  & 22.9 & 1.1 &  55.9 &  2.6\\
German   & 27.4 & 1.3 &  55.4 &  2.5\\
Russian  & 33.3 & 1.3 & 134.9 &  5.3\\
French   & 42.0 & 1.6 &  65.7 &  2.5\\ 
\end{tabular}
\caption{Average number of tokens per sentence (column ``Sent.'') and average
number of tokens per word (column ``Word'') when the training corpus is segmented
by child-specific or parent-specific vocabulary. ``Child-specific'' represents
the effect of using vocabulary customized for examined language. ``EN-CS''
corresponds to the use of English-Czech vocabulary.}
\label{tab:tokenized_dataset}
\end{table}

\begin{figure}[t]
\centering
\begin{tabular}{l|ll}
 & Segmented sentence  \\
\hline               
Original & \foreignlanguage{russian}{Сьерра-Леоне}  \\
EN-RU & \foreignlanguage{russian}{Сьерра\_█-\_█Леоне\_}  \\
EN-CS  & \foreignlanguage{russian}{С█ь█ер█ра█\_█-\_█\textbackslash█10█51█;█е█о█не\_} \\ 
\end{tabular}
\caption{Illustration of segmentation of Russian phrase (gloss: Sierra Leone) with English-Czech and English-Russian vocabulary from our experiments. The character █ represents splits.}
\label{tab:vocab_variant_usage}
\end{figure}

We examine the influence of parent-specific vocabulary on the training dataset of the child. \cref{tab:tokenized_dataset} documents the segmenting effect of different vocabularies. If we compare the child-specific and parent-specific (``EN-CS'') vocabulary, the average number of tokens per sentence or per word increases more than twice.
For example, German has twice as many tokens per word compared to its child-specific vocabulary, and Russian has four times more tokens due to Cyrillic. Odia is affected even more.

Thus, we see that ignoring the vocabulary mismatch introduces a problem for NMT models in the form of an increasing split ratio of tokens. As expected, this is most noticeable for languages using different scripts.

\begin{algorithm}[t]
\begin{center}
\framebox{
\parbox{.9\columnwidth}{
 \SetKwInput{KwData}{Input}
 \KwData{Parent vocabulary (an ordered list of parent subwords) and the training corpus for the child language pair.}
 Generate child-specific vocabulary with the maximum number of subwords equal to the parent vocabulary size\;
 \For{subword S in parent vocabulary}{
  \eIf{S in child vocabulary}{
  continue\;
  }{
  Replace position of S in the parent vocabulary with the first unused child subword not contained in the parent\;
  }
 }
 \KwResult{Transformed parent vocabulary}
}}
 \vspace{1em}
 \caption{Transforming parent vocabulary to contain child subwords and match positions for subwords common for both of language pairs.}
 \label{algorithm}
\end{center}
\end{algorithm}
\subsection{Vocabulary Transformation}
\label{sec:vocabulary_transformation}

Using parent vocabulary roughly doubles the number of subword tokens per word, as we showed in the previous section. This problem would not happen with child-specific vocabulary. However, we are using an already trained parent with its vocabulary. Therefore, we propose a vocabulary transformation method that replaces subwords in the parent wordpiece \parcite{wu2016google} vocabulary with subwords from the child-specific vocabulary.

NMT models associate each vocabulary item with its vector representation
(embedding). When transferring the model from the parent to the child, we decide
which subwords should preserve their embedding as trained in the parent model
and which embeddings should be remapped to new subwords from the child
vocabulary. The goal is to preserve embeddings of subwords that are contained in both parent and child vocabulary. In other words, we reuse embeddings of subwords common to both parent and child vocabularies and reuse the vocabulary entries of subwords not occurring in the child data for other, unrelated, subwords that the child data need. Obviously, the embeddings for these subwords will need to be retrained.




\begin{table*}[t]
\small
\centering
\begin{tabular}{l|r|rrr}
Language pair  & Pairs  & Training set    & Development set    & Test set \\
\hline
EN - Odia & 27k & \perscite{parida2018odiacorpus} & \perscite{parida2018odiacorpus} & \perscite{parida2018odiacorpus} \\   
EN - Estonian & 0.8M & Europarl, Rapid & WMT dev 2018 & WMT 2018 \\    
EN - Finnish & 2.8M & Europarl, Paracrawl, Rapid  & WMT 2015 & WMT 2018\\   
EN - German & 3.5M & Europarl, News commentary, Rapid & WMT 2017 & WMT 2018\\ 
EN - Russian & 12.6M & News Commentary, Yandex, and UN Corpus & WMT 2012 & WMT 2018\\
EN - French & 34.3M & \begin{tabular}{@{}r@{}}Commoncrawl, Europarl, Giga FREN, \\ News commentary,  UN corpus\end{tabular} & WMT 2013 & WMT dis. 2015\\
\end{tabular}
\caption{Corpora used for each language pair. The names specify the corpora from WMT 2018 News Translation Task data. Column ``Pairs'' specify the total number of sentence pairs in training data.}
\label{tab:dataset_sources}
\end{table*}

Our Transformed Vocabulary method starts by constructing the child-specific vocabulary with the size equal to the parent vocabulary size (the parent model is trained, thus it has a fixed number of embeddings). Then, as presented in Algorithm~\ref{algorithm}, we generate an ordered list of child subwords, where subwords known to the parent vocabulary are on the same positions as in the parent vocabulary, and other subwords are assigned arbitrarily to places where parent-only subwords were stored.

We experimented with several possible mappings between the parent and child vocabulary. We tried to assign subwords based on frequency, by random assignment, or based on Levenshtein distance of parent and child subwords. However, all the approaches reached comparable performance; neither of them significantly outperformed the others. One exception is when assigning all subwords randomly, even those that are shared between parent and child. This method leads to worse performance, having several BLEU points lower than other approaches.
Another approach would be to use pretrained subword embeddings similarly as proposed \perscite{kim2019effective}. However, in this paper, we focus on showing, that transfer learning can be as simple as not using any modifications at all.

\section{Experiments}
\label{sec:experiments}

In this section, we first provide the details of the NMT model used in our experiments and the examined set of language pairs. We then discuss the convergence and a stopping criterion and finally present the results of our method for recycling the NMT model as well as improvements thanks to the vocabulary transformation.

\subsection{Parent Model and its Training Data}
\label{sec:model_setting}

In order to document that our method functions in general and is not restricted to our laboratory setting, we do not train the parent model ourselves. Instead, we recycle two systems trained by \perscite{popel-EtAl:2019:WMT}, namely the English-to-Czech and Czech-to-English winning models of WMT 2019 News Translation Task. It is important to note, that we use two parent models and for experiments we always use the parent model with English on the same side, e.g. English-to-Russian child has English-to-Czech as a parent. We leave experimenting with different parents or various combinations for future works, because the goal of this work is to make approach most simple.

We decided to use this model for several reasons. It is trained to translate into Czech, a high-resource language that is dissimilar from any of the languages used in this work.\footnote{The linguistically most similar language of our language selection is Russian, but we do not transliterate Cyrillic into Latin script. Therefore, the system cannot associate similar Russian and Czech words based on appearance.} At the same time, it is trained using the state-of-the-art Transformer architecture as implemented in the Tensor2Tensor framework.\footnote{\url{https://github.com/tensorflow/tensor2tensor}
} \parcite{tensor2tensor}. We use Tensor2Tensor in version 1.8.0.

The model is described in \perscite{popel:2018:WMT}. It is based on the ``Big GPU Transformer'' setup as defined by \perscite{vaswani:2017} with a few modifications. The model uses reverse square root learning rate decay with 8000 warm-up steps and a learning rate of 1. It uses the Adafactor optimizer, the batch size of 2900 subword units, disabled layer dropout.

Due to the memory constraints, we drop training sentences longer than 100 subwords.
We use child hyper-parameter setting equal to the parent model. However, some hyper-parameters like learning rate, dropouts, optimizer, and others could be modified for the training of the child model. We leave these experiments for future work. 

We train models on single GPU GeForce 1080Ti with 11GB memory. In this setup, 10000 training steps take on average approximately one and a half hours. \perscite{popel-EtAl:2019:WMT} trained the model on 8 GPUs for 928k steps, which means that on the single GPU, the parent model would need at least 7424k steps, i.e. more than 45 days of training. 

In our experiments, we train all child models up to 1M steps and then take the model with the best performance on the development set. Because some of the language pairs, especially the low-resource ones, converge within first 100k steps, we use a weak early stopping criterion that stops the training whenever there was no improvement larger than 0.5\% of maximal reached BLEU over the past 50\% of training evaluations (minimum of training steps is 100k). This stopping criterion makes sure that no model is stopped prematurely.

\def\sigboth{\rlap{ $\ddagger$*}}
\begin{table*}[t]
\centering
\begin{tabular}{c|rr|rr|r@{~~~~~~~}rrr}
Language pair & \multicolumn{2}{c}{Baseline} & \multicolumn{2}{|c|}{Direct Transfer} & \multicolumn{4}{c}{Transformed Vocab}\\
                   &  BLEU & Steps & BLEU  &Steps &  BLEU & Steps & $\Delta$ BLEU & Speed-up\\ 
\hline               
English-to-Odia    &  3.54 &   45k &  0.26 &  47k & \bf 6.38\sigboth{} & \bf 38k & 2.84 & 16 \%  \\
English-to-Estonian& 16.03 &   95k & \bf 20.75\significant{} & \bf 75k & 20.27\significant{} & \bf 75k & 4.24 & 21 \%\\
English-to-Finnish & 14.42 &  420k & 16.12\significant{} & \bf 255k & \bf 16.73\sigboth{} & 270k & 2.31 & 36 \% \\
English-to-German  & 36.72 &  270k & 38.58\significant{} & 190k & \bf 39.28\sigboth{} & \bf 110k & 2.56 & 59 \%\\
English-to-Russian & 27.81 & 1090k & 27.04 & 630k & \bf 28.65\sigboth{} & \bf 450k & 0.84 & 59 \% \\
English-to-French  & 33.72 &  820k & 34.41\significant{} & \bf 660k & \bf 34.46\significant{} & 720k & 0.74 & 12 \%\\
\hline
Estonian-to-English & 21.07 & 70k & 24.36\significant{} & \bf 30k  & \bf 24.64\sigboth{}  & 60k & 3.57 & 14 \%\\
Russian-to-English  & 30.31 & 980k & 23.41 & \bf 420k & \bf 31.38\sigboth{}  & 700k & 1.07 & 29 \%\\ 
\end{tabular}
\caption{Translation quality and training time. ``Baseline'' is trained from
scratch with its own vocabulary and child corpus only. ``Direct Transfer'' is
initialized with parent model using the parent vocabulary and continues
training. ``Transformed Vocab'' has the same initialization but merges the
parent and child vocabulary as described in
\cref{sec:vocabulary_transformation}. Best score and lowest training time in
each row in bold. The statistical significance is computed against the baseline (\significantmark{}) or against ``Direct Transfer'' (*).
Last two columns show improvements of Transformed Vocabulary in comparison to the baseline.
}
\label{tab:main_results}
\end{table*}

\subsection{Studied Language Pairs}
\label{sec:studied_languages}

We use several child language pairs to show that our approach is useful for various sizes of corpora, language pairs, and scripts. To cover this range of situations, we select languages in \cref{tab:dataset_sources}. Future works could focus also on languages outside from Indo-European family, such as Chinese.

Another decision behind selecting these language pairs is to include language pairs reaching various levels of translation quality. This is indicated by automatic scores of the baseline setups ranging from 3.54 BLEU (English-to-Odia) to 36 BLEU (English-to-German)\footnote{The systems submitted to WMT 2018 for English-to-German translation have better performance than our baseline due to the fact, that we decided not to use Commoncrawl, which artificially made English-German parallel data less resourceful.}, see \cref{tab:main_results}.

The sizes of corpora are in \cref{tab:dataset_sources}. The smallest language pair is English-Odia, which uses the Brahmic writing script and contains only 27 thousand training pairs. The largest is the high-resource English-French language pair.

For most of the language pairs, we use training data from WMT \parcite{wmt_2018}.\footnote{\url{http://www.statmt.org/wmt18/}} We use the training data without any preprocessing, not even tokenization.\footnote{While the recommended best practice in past WMT evaluations was to use Moses tokenizer. It is not recommended for Tensor2Tensor with its build-in tokenizer any more.} See \cref{tab:dataset_sources} for the list of used corpora for each language pair. For some languages, we have opted out from using all available corpora in order to experiment on languages containing various magnitudes of parallel sentences.

For high-resource English-French language pair, we perform a corpora cleaning using language detection Langid.py \cite{lui-baldwin:2012:Demo}.
We drop all sentences that are not recognized as the correct language. It removes 6.5M (15.9
\%) sentence pairs from the English-French training corpora.

\section{Results}
\label{sec:results}

All reported results are calculated on the test data and evaluated with SacreBLEU \parcite{sacrebleu}. The results are in \cref{tab:main_results}. We discuss separately the training time, automatically assessed translation quality using the parent and the Transformed Vocabulary, and comparison to \perscite{kocmi-adaption} in the following sections.

Baselines use the same architecture, and they are trained solely on the child training data with the use of child-specific vocabulary. We compute statistical significance with a paired bootstrap resampling \parcite{bootstrap-koehn:2004}. We use 1000 samples and a confidence level of 0.05. Statistically significant improvements are marked by \significantmark{}.

\subsection{Direct Transfer Learning} 

First, we compare the Direct Transfer learning in contrast to the baseline. We see that Direct Transfer learning is significantly better than the baseline in both translation directions in all cases except for Odia and Russian, which we will discuss later. We get improvements for various language types, as discussed in \cref{sec:studied_languages}. The largest improvement is of 4.72 BLEU for the low-resource language pair of Estonian-English, but we also get an improvement of 0.69 BLEU for the high-resource pair French-English. 

The results are even more surprising when we take into account the fact that the model uses the parent vocabulary, and it is thus segmenting words into considerably more subwords. This suggests that the Transformer architecture generalizes very well to short subwords.
However, the worse performance of English-Odia and English-Russian can be attributed to the different writing script. The Odia script is not contained in the parent vocabulary at all, leading to segmenting of each word into individual bytes, the only common units with the parent vocabulary. 
Therefore, to avoid problems with filtering, we increase the filtering limit of long sentences during training from 100 to 500 subwords for these two language pairs (see \cref{sec:model_setting}).

\subsection{Results with Transformed Vocabulary}

As the results in \cref{tab:main_results} confirm, Transformed Vocabulary successfully tackles the problem of the child language using a different writing script.
We see ``Transformed Vocab'' delivering the best performance for all language
pairs except for English-to-Estonian, significantly improving over baseline and even
over ``Direct Transfer'' in most cases.

\subsection{Training Time}

In the introduction, we discussed that recent development in NMT focuses mainly on the performance over efficiency \parcite{schwartz2019green}. Therefore, in this section, we discuss the amount of training time required for our method to converge. We are reporting the number of updates (i.e. steps) needed to get the model used for evaluation.\footnote{Another possibility would be to report wall-clock time. However, that is influenced by server load and other factors. The number of steps is better for the comparison as long as the batch size stays the same across experiments.}

We see in \cref{tab:main_results} that both our methods converged in a lower number of steps than the baseline. For the Transformed Vocabulary method, we get a speed-up of 12--59 \%. The reduction in the number of steps is most visible in English-to-German and English-to-Russian. 
It is important to note that the number of steps to the convergence is not precisely comparable, and some tolerance must be taken into account. It is due to the fluctuation in the training process. However, in neither of our experiments, Transformed Vocabulary is slower than baseline. Thus we conclude that our Transformed Vocabulary method takes fewer training steps to finish training than training a model from scratch.

\begin{table}[t]
\centering
\small
\begin{tabular}{ll|rrr}
&Language &  & Transf. &     Warm\\
&pair     & Baseline & vocab & Start \\
\hline               
\multirow{4}{*}{\rot{BLEU}}&To Estonian   & 16.03  & 20.27 & \bf 20.75 \\ 
&To Russian    & 27.81 & 28.65 &\bf 29.03\significant{}  \\
&From Estonian & 21.07 & 24.64 &\bf 26.00\significant{}  \\
&From Russian  & 30.31 & \bf 31.38  & 31.15   \\
\hline
\multirow{4}{*}{\rot{Steps}}&To Estonian   & 95k  & \bf 75k & 735k  \\
&To Russian    & 1090k  & \bf 450k& 1510k \\
&From Estonian & 70k & \bf 60k& 700k   \\
&From Russian  & 980k& \bf 700k & 1465k  \\
\end{tabular}
\caption{Comparison of our Transformed Vocabulary method with \perscite{kocmi-adaption} (abridged as ``Warm Start''). The top half of table compares results in BLEU, the bottom half the number of steps needed to convergence. Steps of \perscite{kocmi-adaption} method are reported as the sum of parent and child training, due to the  nature of the method.
}
\label{tab:transfer_compar}
\end{table}

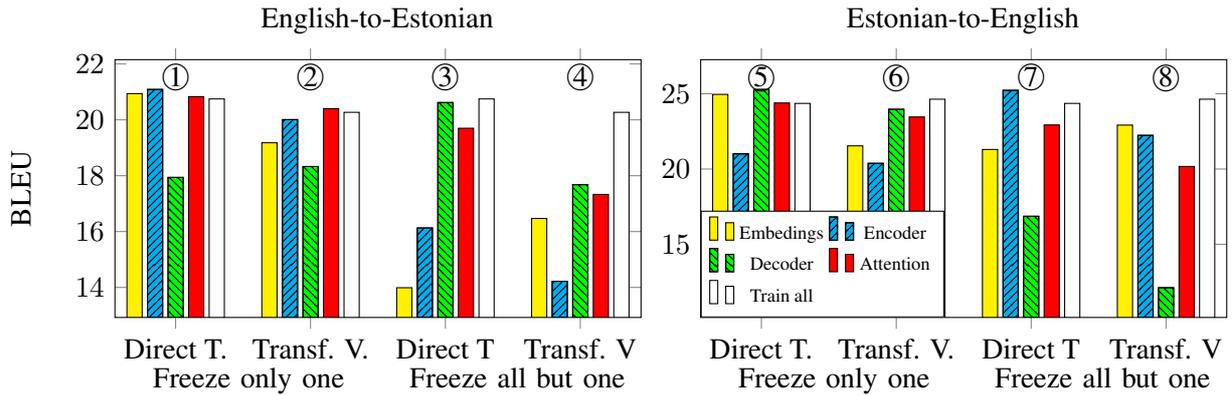
\begin{figure*}
\begin{tikzpicture}
\begin{axis}[
    ybar,
    bar width=.2cm,
    width=8.5cm,
    height=5cm,
    enlargelimits=0.15,
    legend style={at={(1,0)},
      anchor=south east,font=\scriptsize,legend columns=2},
    ylabel={BLEU},
    symbolic x coords={Direct T., Transf. V., Direct T, Transf. V},
    xtick=data,
    title=English-to-Estonian,
    xlabel style={text width=8.5cm}, 
    xlabel=\hspace{33pt} Freeze only one \hspace{17pt} Freeze all but one
    ]
\addplot +[black, fill=yellow] coordinates {(Direct T.,20.94)(Transf. V.,19.18)(Direct T,13.99)(Transf. V,16.47)};
\addplot +[black, fill=cyan, postaction={pattern=north east lines}] coordinates {(Direct T.,21.09)(Transf. V.,20.01)(Direct T,16.13)(Transf. V,14.22)};
\addplot +[black, fill=green, postaction={pattern=north west lines}] coordinates {(Direct T.,17.94)(Transf. V.,18.33)(Direct T,20.62)(Transf. V,17.68)};
\addplot +[black, fill=red] coordinates {(Direct T.,20.83)(Transf. V.,20.40)(Direct T,19.70)(Transf. V,17.33)};
\addplot +[black, fill=white] coordinates {(Direct T.,20.75)(Transf. V.,20.27)(Direct T,20.75)(Transf. V,20.27)};
\node[] at (axis cs: Direct T.,21.5) {\circled{1}};
\node[] at (axis cs: Transf. V.,21.5) {\circled{2}};
\node[] at (axis cs: Direct T,21.5) {\circled{3}};
\node[] at (axis cs: Transf. V,21.5) {\circled{4}};
\end{axis}

\begin{scope}[shift={(7.7,0)}]
\begin{axis}[
    ybar,
    bar width=.2cm,
    width=8.5cm,
    height=5cm,
    enlargelimits=0.15,
    legend style={at={(0,0)},
      anchor=south west,font=\scriptsize,legend columns=2},
    symbolic x coords={Direct T., Transf. V., Direct T, Transf. V},
    xtick=data,
    title=Estonian-to-English,
    xlabel style={text width=8.5cm}, 
    xlabel=\hspace{33pt} Freeze only one \hspace{17pt} Freeze all but one
    ]
\legend{Embedings,Encoder,Decoder,Attention,Train all}
\addplot +[black, fill=yellow] coordinates {(Direct T,21.30)(Transf. V,22.92)(Direct T.,24.96)(Transf. V.,21.54)};
\addplot +[black, fill=cyan, postaction={pattern=north east lines}] coordinates {(Direct T,25.23)(Transf. V,22.24)(Direct T.,21.01)(Transf. V.,20.38)};
\addplot +[black, fill=green, postaction={pattern=north west lines}] coordinates {(Direct T,16.86)(Transf. V,12.11)(Direct T.,25.29)(Transf. V.,23.97)};
\addplot +[black, fill=red] coordinates {(Direct T,22.93)(Transf. V,20.17)(Direct T.,24.39)(Transf. V.,23.46)};
\addplot +[black, fill=white] coordinates {(Direct T,24.36)(Transf. V,24.64)(Direct T.,24.36)(Transf. V.,24.64)};
\node[] at (axis cs: Direct T.,26) {\circled{5}};
\node[] at (axis cs: Transf. V.,26) {\circled{6}};
\node[] at (axis cs: Direct T,26) {\circled{7}};
\node[] at (axis cs: Transf. V,26) {\circled{8}};
\end{axis}
\end{scope}
\end{tikzpicture}
\caption{Child BLEU scores when trained with some parameters frozen. The left plot shows English-to-Estonian and the right is Estonian-to-English. In both plots, the first two groups are experiments where one component is frozen and the second two are when all components but one are frozen.}
\label{tab:freezing_analysis}
\end{figure*}

\subsection{Comparison to \perscite{kocmi-adaption}}

We replicated the experiments of \perscite{kocmi-adaption} with the identical
framework and hyperparameter setting in order to compare their method to ours.
We experiment with Estonian-English and Russian-English language pair in both
translation directions. Their approach needs an individual parent for every
child model, so we train four models: two English-to-Czech and two
Czech-to-English on the same parent training data as \perscite{kocmi-adaption}.
All vocabularies contain 32k subwords.
We compare their method with our Transformed Vocabulary. Furthermore, the results of Direct Transfer in \cref{tab:main_results} are also comparable with this experiment.

In \cref{tab:transfer_compar}, we see that their method reaches a slightly better performance in three translation models, where English-to-Russian and Estonian-to-English are significantly (\significantmark{}) better than Transformed Vocabulary technique; the other two are on par with our method, which is understandable. The Transformed Vocabulary cannot outperform the warm-start technique since the warm-start parent model has the advantage of being trained with the vocabulary prepared for the investigated child.

However, when we compare the total number of steps needed to reach the performance, both our approaches are significantly faster than \perscite{kocmi-adaption}. The most substantial improvements are roughly ten times faster for Estonian-to-English, and the smallest difference for English-to-Russian is two times faster.
This is mostly because their method first needs to train the parent model that is specialized for the child, while our method can directly re-use any already trained model. Moreover, we can see that their method is even slower than the baseline model.

\section{Analysis by Freezing Parameters}
\label{sec:analysis}


To discover which transferred parameters are the most helpful for the child
model and which need to be changed the most, we follow the analysis used by
\perscite{thompson-EtAl:2018:WMT}: When training the child, we freeze some of
the parameters.

Based on the internal layout of the Transformer model in Tensor2Tensor, we divide the model into four components. (i) Word embeddings (shared between encoder and decoder) map each subword unit to a dense vector representation. 
(ii) The encoder component includes all the six feed-forward layers converting the input sequence to the deeper representation. (iii) The decoder component consists again of six feed-forward layers preparing the choice of the next output subword unit. (iv) The multi-head attention is used throughout encoder and decoder, as self-attention layers interweaved with the feed-forward layers. 

We run two sets of experiments: either freeze only one out of the four components and leave the rest of the model updating or freeze everything but the examined component. We also test it on two translation directions: \langpair{English}{Estonian} in the left hand part of \cref{tab:freezing_analysis} and \langpair{Estonian}{English} in the right hand part. In both cases, English-Czech (in the corresponding direction, i.e. with English on the correct side) serves as the parent.
We discuss individual components separately, indexing the experiments
\circled{1} to \circled{8}.



Similarly to \perscite{thompson-EtAl:2018:WMT} in domain adaptation, we observe
that parent embeddings serve well in Direct Transfer, freezing them has a
minimal impact compared to the baseline in \circled{1} and \circled{5}.
The frozen embeddings in Transformed
Vocabulary (\circled{2}, \circled{6}) results in significant performance drops
which can be attributed to the arbitrary assignment of embeddings to new subwords.

The comparison of all but embeddings frozen in \circled{4} and
\circled{8} (Transformed Vocabulary) is interesting. In \circled{8},
the performance of the network can be recovered close to the
baseline by retraining either parent source embeddings or the encoder. These
two components can compensate for each other.
%
%
This differs from
the case with English reused in the source (\circled{4}) where updating
embeddings to the child language is insufficient: the decoder must be updated to produce
fluent output in the new target language and even with the decoder updated, the
loss compared to the baseline is quite substantial.


The most important component for transfer learning is generally the
component handling the new language:
decoder in English-to-Estonian
and encoder in
the reverse. 
With this component fixed, 
the performance drops the most with this component fixed (\circled{1},
\circled{2}, \circled{5}, \circled{6}) and among the least with this component free to
update (\circled{3}, \circled{4}, \circled{7}, \circled{8}).
%
%
This confirms that at least for examined language pair, the Transformer model
lends itself very well to encoder or decoder re-use.

Other results in \cref{tab:freezing_analysis} reveal that the architecture can compensate for some of the training deficiencies. Freezing the encoder \circled{1}, \circled{2} (resp. decoder for Estonian-to-English \circled{5}, \circled{6}) or attention is not that critical as the frozen decoder (resp. encoder). The bad result of the encoder \circled{3}, \circled{4} (resp. decoder \circled{7}, \circled{8}) being the only non-frozen component shows that model is not capable of providing all the needed capacity for the new language, unlike the self-attention where the loss is not that large. This behaviour correlates with our intuition that the model needs to update the most the component that handles the differing language with the parent model (in our case Czech).

All in all, these experiments illustrate the robustness of the Transformer model that it is able to train and reasonably well utilize pre-trained weights even if they are severely crippled.

\section{Related Work}
\label{sec:related}

This paper focuses on re-using an existing NMT model in order to improve the performance in terms of training time and translation quality without any need to modify the model or pre-trained weights. 

\perscite{dynamic_vocabulary} presented two model modifications for multilingual MT and showed that transfer learning could be extended to transferring from the parent to the first child, followed by the second child and then the third one. They achieved improvements with dynamically updating embeddings for the vocabulary of a target language.

The use of other language pairs for improving results for the target language pair has been approached from various angles. One option is to build multilingual models \parcite{liu2020multilingual}, ideally so that they are capable of zero-shot, i.e. translating in a translation direction that was never part of the training data.
\perscite{TACL1081} and \perscite{lu-EtAl:2018:WMT1} achieve this with a unique language tag that specifies the desired target language.
The training data includes sentence pairs from multiple language pairs, and the model implicitly learns translation among many languages. In some cases, it achieves zero-shot and can translate between languages never seen together. \perscite{gu-EtAl:2018:N18-1} tackled the problem by creating universal embedding space across multiple languages and training many-to-one MT system. \perscite{firat-cho-bengio:etal:2016} propose multi-way multi-lingual systems. Their goal is to reduce the total number of parameters needed to train multiple source and target models. In all cases, the methods are dependent on a special training schedule.

The lack of parallel data in low-resource language pairs can also be tackled by unsupervised translation \parcite{artetxe2017unsupervised,lample2018phrase}. The general idea is to train monolingual autoencoders for both source and target languages separately, followed by mapping both embeddings to the same space and training simultaneously two models, each translating in a different direction. In an iterative training, this pair of NMT systems is further refined, each system providing training data for the other one by back-translating monolingual data \parcite{sennrich-haddow-birch:2016:monolingual}.

For very closely related language pairs, transliteration can be used to generate training data from a high-resourced pair to support the low-resourced one as described in \perscite{Karakanta2018}.

\section{Conclusion}
\label{sec:conclusion}

In this paper, we focus on a setting where existing models are re-used without any preparation for knowledge transfer of original model ahead of its training. This is a relevant and prevailing situation in academia due to computing restrictions, and industry, where updating existing models and scaling to more language pairs is essential.
We evaluate and propose methods of re-using Transformer NMT models for any ``child'' language pair regardless of the original ``parent'' training languages and especially showing, that no modification is better than training from scratch. 

The techniques are simple, effective, and applicable to models trained by others which makes it more likely that our experimental results will be replicated in practice. 
We showed that despite the random assignment of subwords, the Transformed Vocabulary improves the performance and shortens the training time of the child model compared to training from random initialization.

Furthermore, we showed that this approach is not restricted to low-resource languages, and we documented that the highest improvements are (expectably) due to the shared English knowledge. 
Moreover, we confirmed the robustness of the Transformer and its ability to achieve good results in adverse conditions like very fragmented subword units or parts of the network frozen.

The warm-start approach by \perscite{kocmi-adaption} performs slightly better than our Transformed Vocabulary, but it needs to be trained for a significantly longer time. 
This leaves room for approaches that also focus on the efficiency of the training process. We perceive our approach as a technique for increasing the performance of a model without an increase in training time. Thus, re-using older models in cold-start scenario of transfer learning can be used in standard NMT training pipelines without any performance or speed losses instead of random initialization as is the common practice currently.

\section*{Acknowledgements}

This study was supported in parts by the grants 18-24210S of the Czech Science Foundation and 825303 (Bergamot) of the European Union.
This work has been using language resources and tools stored and distributed by the LINDAT/CLARIN project of the Ministry of Education, Youth and Sports of the Czech Republic (LM2015071).

\bibliography{biblio}
\bibliographystyle{eamt20}

\end{document}